\documentclass[12pt]{l4dc2020}

\usepackage{times}
\usepackage{appendix}
\usepackage[noabbrev,capitalise,nameinlink]{cleveref}

\DeclareMathOperator*{\argmax}{arg\,max}

\allowdisplaybreaks

\title[Periodic Q-Learning]{Periodic Q-Learning}

\author{%
 \Name{Donghwan Lee} \Email{donghwan@kaist.ac.kr}\\
 \addr Department of Electrical Engineering, Korea Advanced Institute of Science and Technology
 \AND
 \Name{Niao He} \Email{niaohe@illinois.edu}\\
 \addr Department of Industrial and Enterprise Systems Engineering, University of Illinois at Urbana-Champaign%
}

\begin{document}

\maketitle

\begin{abstract}%
 The use of target networks is a common practice in deep reinforcement learning for stabilizing the training; however, theoretical understanding of this technique is still limited. In this paper, we study the so-called \emph{periodic Q-learning} algorithm (PQ-learning for short), which resembles the technique used in deep Q-learning for solving infinite-horizon discounted Markov decision processes (DMDP) in the tabular setting. PQ-learning maintains two separate Q-value estimates -- the online estimate and target estimate. The online estimate follows the standard Q-learning update, while the target estimate is updated \emph{periodically}.  In contrast to the standard Q-learning, PQ-learning enjoys a simple finite time analysis and achieves better sample complexity for finding an $\varepsilon$-optimal policy. Our result provides a preliminary justification of the effectiveness of utilizing target estimates or networks in Q-learning algorithms.

\end{abstract}


\section{Introduction}
Reinforcement learning (RL) addresses the optimal control problem for unknown systems through experiences~\citep{sutton1998reinforcement}. Among many others, Q-learning~\citep{watkins1992q} is one of the most popular RL algorithms. Recent deep Q-learning~\citep{mnih2015human} has captured significant attentions in the RL community for outperforming humans in several challenging tasks. Besides the effective use of deep neural networks as function approximators, the success of deep Q-learning is also indispensable to the utilization of target networks when calculating target values at each iteration. Specifically, deep Q-learning maintains two separate networks, the Q-network that approximates the state-action value function, and the target network that is synchronized with the Q-network periodically.
In practice, using target networks is proven to substantially improve the performance of Q-learning algorithms~\citep{mnih2015human}. However, theoretical understanding of this technique remains rather limited. \citet{lee2019target} recently explores a family of target-based temporal-difference learning algorithms for policy evaluation and develops their convergence analyses. \cite{yang2019theoretical} provides theoretical analysis of the neural fitted Q-iteration under some simplification.

In this paper, we investigate a simple algorithm, called the \emph{periodic Q-learning} (PQ-learning), for finding the optimal policy of infinite-horizon discounted Markov decision process (DMDP) in the tabular setting (i.e., discrete finite state-space). The algorithm mimics deep Q-learning by maintaining two separate Q-value estimates -- the online estimate and target estimate: the online estimate takes the standard Q-learning updating rule while freezing the target estimate, and the target estimate is updated through a periodic fashion.
Formally speaking, consider the DMDP represented by the tuple $(\mathcal{S},\mathcal{A}, P, r,\gamma)$, where $\mathcal{S}$ and $\mathcal{A}$ are the finite state and action spaces, $r(s,a)\in[-1,1]$ is the state-action reward,  $P_a(s,s')$ is the probability of transiting from $s$ to $s'$ when taking action $a$, and $\gamma\in(0,1)$ is a discount factor. The objective is to find a deterministic optimal policy, $\pi^*:\mathcal{S}\to\mathcal{A}$, such that the cumulative discounted rewards over infinite time horizons is maximized, i.e., $\pi^*:=\argmax_{\pi} {\mathbb E}\left[\left.\sum_{k=0}^\infty {\gamma^k r(s_k,\pi(s_k))}\right|\pi\right].$ For simplicity, throughout, we assume that we have access to a sequence of \emph{i.i.d.} random variables $\{(s_k,a_k)\}_{k=0}^{\infty}$ from a fixed underlying probability distribution, $d_a(s),s\in {\cal S},a \in {\cal A}$, of the state and action pair $(s,a)$.\footnote{This assumption can be relaxed to ergodicity conditions. But for simplicity, we focus only on the \emph{i.i.d.} case.} The PQ-learning algorithm is formally presented in Algorithm~\ref{algo:PQ-learning-short}.

\begin{algorithm}[t]
\SetAlgoLined
\hrulefill\\
Initialize $Q_0(s,a)\in[-\frac{1}{1-\gamma},\frac{1}{1-\gamma}], \forall(s,a)\in {\cal S} \times {\cal A}$; iteration  $T$; steps $\{N_k\}_{k=0}^{T-1}$; stepsizes $\{\beta_t\}_{t=0}^\infty$;
\For{iteration $k=0,1,\ldots, T-1$}{
\For{step $t=0,1,\ldots, N_k-1$}{

Obtain sample $(s,a)\sim d_a(s)$, $s'\sim P_a(s,\cdot)$ and $r(s,a)$\;

Update $Q_{k,t+1}(s,a)=Q_{k,t}(s,a) - \beta_t (r(s,a)+{\gamma}\max_{a'\in {\cal A}}Q_k(s',a')-Q_{k,t}(s,a))$\;
}
Update $Q_{k+1}=Q_{k,N_k}$\;
}
Return $Q_T$\;

\hrulefill
\caption{Periodic Q-Learning (PQ-learning)}
\end{algorithm}\label{algo:PQ-learning-short}

Notice that when setting $N_k=1$ for all $k$, PQ-learning reduces to the standard Q-learning. To the authors' knowledge, finite-time convergence analysis of the standard Q-learning for DMDPs can be challenging, and only few results are reported in the literature. In particular,~\citet{even2003learning} shows that to achieve an $\varepsilon$-optimal Q-function such that $\|Q_T-Q^*\|_2\leq\varepsilon$ with probability at least $1-\delta$, the number of samples required by the standard Q-learning algorithm with linear learning rate is
${\cal O} \Big( \frac{(|{\cal S}||{\cal A}|)^{\frac{2\ln(1/\varepsilon)}{1-\gamma}}}{(1-\gamma)^2\varepsilon^2}\ln\Big(\frac{|{\cal S}||{\cal A}|}{\delta(1-\gamma)\varepsilon}\Big)\Big).$
However, for PQ-learning, we show that the finite sample analysis can be easily established based on standard tools of analyzing the stochastic gradient descent and the contraction property of Bellman equation.  Our analysis hinges on a key observation that PQ-learning can be viewed as solving a sequence of mean-squared Bellman error minimization subproblems through the stochastic gradient descent routine.

As a main result, we prove that to find an $\varepsilon$-optimal Q function such that ${\mathbb E}[\|Q_T-Q^*\|_\infty]\le \varepsilon$, the number of samples needed for  PQ-learning is at most
\begin{align*}
{\cal O}\left(\frac{|{\cal S}||{\cal A}|}{\varepsilon^2 (1-\gamma)^4}\frac{L}{c^3}\ln \left( {\frac{1}{(1-\gamma)^2\varepsilon}}\right)\right),
\end{align*}
where $c$ and $L$ corresponds to the minimal and maximal probabilities of the state-action distribution $d_a(s)$. The most efficient sample complexity can be achieved when the state-action distribution is uniform. In this case, the sample complexity becomes
\begin{align*}
{\cal O}\left(\frac{|{\cal S}|^3|{\cal A}|^3}{\varepsilon^2 (1-\gamma)^4}\ln \left( {\frac{1}{(1-\gamma)^2\varepsilon}}\right)\right).
\end{align*}

\begin{table}[t]
\caption{Comparison of complexities in terms of $|{\cal S}|$, $|{\cal A}|$, and $\varepsilon$ up to logarithmic factors. The notation ``$-$'' means that the run-time complexity is the same order as the sample complexity.}\label{table:comparison}
\vspace{-0.25cm}
\begin{center}
\begin{tabular}{c c c}
\hline
Algorithm & Sample Complexity  & Run-time Complexity\\
\hline
\textbf{PQ-learning} (this paper) & $\tilde{\cal O}\left(\frac{|{\cal S}|^3|{\cal A}|^3}{\varepsilon^2}\right)$  & $-$ \\
\hline
Q-learning~\citep{even2003learning} & $\tilde{\cal O} \Big( \frac{(|{\cal S}||{\cal A}|)^{\frac{2\ln(1/\varepsilon)}{1-\gamma}}}{\varepsilon^2}\Big)$ & $-$  \\

\hline
Delayed Q-learning~\citep{strehl2006pac} & $\tilde{\cal O}\left(\frac{|{\cal S}||{\cal A}|}{\varepsilon^4}\right)$  & $-$ \\
\hline
Phased Q-learning~\citep{kearns1999finite} & $\tilde{\cal O}\left(\frac{|{\cal S}||{\cal A}|}{\varepsilon^2}\right)$  & N/A \\
\hline
SPD Q-learning~\citep{lee2018stochastic} & $\tilde{\cal O}\left(\frac{|{\cal S}|^8|{\cal A}|^8}{\varepsilon^2}\right)$ & $-$ \\
\hline
Lower bound~\citep{azar2013minimax} & $\Omega\left(\frac{|{\cal S}||{\cal A}|}{\varepsilon^2} \right)$  & N/A\\ 
\hline
\end{tabular}
\end{center}
\end{table}

Compared to the standard Q-learning, PQ-learning is also fairly simple to implement, with $\mathcal{O}(1)$  per-iteration complexity. On the theory side, the sample complexity (also run-time complexity) of PQ-learning greatly improves that of the standard Q-learning, in terms of the dependence on the sizes of state and action spaces. Our result, in some sense, sheds light on the effectiveness of using target estimates or networks commonly observed in practice. Finally, while the current paper only focuses on the tabular setting, PQ-learning can be extended to incorporate linear function approximation or neural network approximation (i.e., similar to deep Q-learning). We hope this work would open the door to further investigation of the theory of deep Q-learning  and the design of more efficient reinforcement learning algorithms with target networks. 

\paragraph{Related works.} There exists a significant body of Q-learning variations for tabular DMDP in the literature. Some representative model-free algorithms include phased Q-learning~\citep{kearns1999finite},  delayed Q-learning~\citep{strehl2006pac},  fitted Q-learning~\citep{ernst2005tree},  double Q-learning~\citep{hasselt2010double},  Zap Q-learning~\citep{devraj2017zap}, stochastic primal-dual Q-learning (SPD Q-learning)~\citep{lee2018stochastic}, etc. Besides, there are other sampling algorithms for tabular DMDP, such as R-MAX~\citep{strehl2009reinforcement}, empirical QVI~\citep{azar2013minimax}, sublinear randomized QVI~\citep{sidford2018near}, etc. We briefly summarize the computational complexities of some representative works in Table~\ref{table:comparison}, ignoring the dependence on the logarithmic factors and the discount factor.\footnote{The complexities in Table~\ref{table:comparison} are obtained after simplifications for convenience of presentations, to provide a rough overview rather than a detailed comparative analysis since many algorithms operate under different assumptions.} We can see that PQ-learning is reasonably efficient comparing to several existing  Q-learning variations. For instance, the phased Q-learning \citep{kearns1999finite} has lower sample complexity, but its run-time complexity  can be very large. The delayed Q-learning~\citep{strehl2006pac} is efficient in that the complexity scales linearly in $|{\cal S}|$ and $|{\cal A}|$, but the complexity in terms of $\varepsilon$ is much worse than PQ-learning. Note that the sample complexity achieved by PQ-learning, is by no means optimal, in terms of the dependence on $|\mathcal{S}|,|\mathcal{A}|$ and the factor $1-\gamma$, comparing to the lower bound established in the literature for DMDP under a generative sampling model~\citep{strehl2009reinforcement,azar2013minimax}. However, we emphasize that the goal of this work is to understand the convergence and efficiency of target-based Q-learning, rather than establishing an optimal algorithm that matches the sample complexity lower bound for tabular DMDP.

\section{Periodic Q-learning}
Recall that the Q-function under policy $\pi$ is defined as
$Q^{\pi}(s,a)={\mathbb E}\big[\left. \sum_{k=0}^\infty {\gamma^k r(s_k,\pi(s_k))}\right|s_0=s,a_0=a,\pi\big],s\in {\cal S},a\in {\cal A},$
and the optimal Q-function is defined as $Q^*(s,a)=Q^{\pi^*}(s,a)$ for all $s\in {\cal S},a\in {\cal A}$. Consider the Bellman operator $\mathbf{T}: {\mathbb R}^{|{\cal S}||{\cal A}|} \to {\mathbb R}^{|{\cal S}||{\cal A}|}$,
\begin{align*}
&(\mathbf{T}Q)(s,a):=\sum_{s'\in {\cal S}}{P_a (s,s')\left(r(s,a)+\gamma \max _{a' \in A} Q(s',a')\right)}.
\end{align*}
The Bellman operator is known to be a contraction with respect to the max-norm~\citep{puterman2014markov}, and the optimal Q-function, $Q^*$, is the unique fixed point of this operator. Once $Q^*$ is known, then an optimal policy can be retrieved by $\pi^*(s)=\argmax_{a\in {\cal A}}Q^*(s,a)$. Therefore, the MDP can be solved by finding the optimal Q-function.


For the PQ-learning, at each iteration $k$, the algorithm can be viewed as approximately computing the Bellman operator $\mathbf{T}Q_k$ through minimizing the mean-squared loss function
\begin{align}
\min_{Q\in {\mathbb R}^{|{\cal S}||{\cal A}|}} l(Q;Q_k):=\frac{1}{2} {\mathbb E}_{s,a}\left[\left( {\mathbb E}_{s'}\left[r(s,a)+\gamma\max_{a'\in {\cal A}}Q_k(s',a')\right]-Q(s,a)\right)^2 \right].\label{eq:Q-learning-objective}
\end{align}
Here ${\mathbb E}_{s,a}$ is the expectation taken with respect to the current state-action pair which has the distribution $d_a(s)$, i.e., ${\mathbb P}[s_k=s,a_k=a]=d_a(s)$ for any time step $k\geq 0$, and ${\mathbb E}_{s'}$ is the expectation taken with respect to the next state $s'\sim P_a(s,\cdot)$. In particular, PQ-learning approximately solves the subproblem (\ref{eq:Q-learning-objective}) through $N_k$ steps of stochastic gradient descent:
\begin{align*}
&Q_{k,t+1}=Q_{k,t}-\beta_t \left. \tilde\nabla_Q l(Q;Q_k) \right|_{Q=Q_{k,t}}, \;t=0,\ldots, N_k-1
\end{align*}
where $\tilde \nabla_Q l(Q;Q_{k})$ is a stochastic estimator of the gradient $\nabla_Q l(Q;Q_{k})$. Here, $Q_{k}$ can be treated as the target estimate of the optimal Q-values, and $Q_{k,t}$ is the online estimate. The target estimate is synchronized with the online estimate after $N_k$ step, i.e., $Q_{k+1}=Q_{k,N_k}$. Here, for generality, we allow the inner steps $N_k$ to vary at every iteration. PQ-learning resembles the original deep Q-learning~\citep{mnih2015human} when $N_k$ is set to a constant.  If $N_k=0$ for all $k = 0,1,\ldots,T-1$, then PQ-learning corresponds to the standard Q-learning.

As we only apply a finite number $N_k$ steps of SGD to solve (\ref{eq:Q-learning-objective}), the SGD subroutine will return an approximate solution with a certain error bound. Throughout, we denote $\{\varepsilon_k\}_{k=1}^{T}$ as the approximation errors such that
$${\mathbb E}[\| Q_{k+1}-\mathbf{T}Q_{k}\|_2^2]\le\varepsilon_{k+1}.$$
As a result, the convergence of PQ-learning depends on the cumulative approximation errors $\{\varepsilon_k\}_{k=1}^{T}$  induced from SGD subroutine, whereas these error terms are determined by the number of inner steps, the learning rate, and the variance of stochastic gradient. In what follows, we derive the convergences of the outer and inner iterations, and further establish the overall sample complexity for achieving $\varepsilon$-optimal policy.

\paragraph{Remark.} It is worth pointing out that PQ-learning shares some similarity with fitted Q-iteration (FQI)~\citep{ernst2005tree} algorithms in that both approaches can be viewed as approximating the Bellman operator at every iteration, but there exist notable differences. FQI usually solves the resulting linear regression problem through matrix inversion computation or batch algorithms, requiring expensive memory and computation cost. On the other hand, PQ-learning solves the subproblems through the online SGD subroutine, which is much more efficient.  By using the SGD steps, PQ-learning can be naturally interpreted as Q-learning with periodic target update (Algorithm 1), thus much simpler than the FQI algorithms. As will be shown in the next section, PQ-learning enjoys a much simpler finite-time convergence and complexity analysis, in stark contrast to many other online Q-learning algorithms. 

\section{Main Results}
Throughout, we assume that the sampling distribution satisfies that $d_a(s)> 0$ for all $s\in {\cal S},a \in {\cal A}$. Throughout the paper, we define the constants $c:=\min_{s\in {\cal S},a \in {\cal A}} d_a(s)$ and $L:=\max_{s\in {\cal S},a \in {\cal A}} d_a(s)$, the nonsingular diagonal matrix $D:=\text{diag}[D_1;\ldots, D_{|{\cal A}|}]\in {\mathbb R}^{|{\cal S}||{\cal A}| \times |{\cal S}||{\cal A}|}$, where each $D_a, a\in\{1,2,\cdots,|\cal A|\}$ is a diagonal matrix whose diagonal entries consist of the distribution $d_a(s), s\in {\cal S}$. We start by characterizing the outer and inner iteration convergence, respectively.
\begin{proposition}[Outer iteration convergence]\label{prop:outer-iteration}
We have
\begin{align*}
&{\mathbb E}[\|Q_T-Q^*\|_\infty]\le\sum_{k=1}^T {\gamma^{T-k}\sqrt{\varepsilon_k}}+\gamma^T {\mathbb E}[\|Q_0-Q^*\|_\infty].
\end{align*}
Particularly, if $\varepsilon_k=\varepsilon$ for all $k \geq 0$, then
${\mathbb E}[\|Q_T-Q^*\|_\infty] \le \frac{\sqrt\varepsilon}{1-\gamma} + \gamma^T {\mathbb E}[\|Q_0-Q^*\|_\infty].$
\end{proposition}

One can see that the error is essentially decomposed into two terms, one from the approximation errors induced from SGD procedures and one from the contraction property of solving the subproblems, which can also be viewed as approximately computing the Bellman operators. To further analyze the approximation error from the SGD procedure,  existing convergence results for SGD can be applied with some modifications.

\begin{proposition}[Inner iteration convergence]\label{prop:inner-iteration}
Suppose that ${\mathbb E}[\|Q_i-\mathbf{T}Q_{i-1}\|_2^2]\le\varepsilon,i\in \{1,2,\ldots ,k\}$ and $\varepsilon\le (1-\gamma)^2$ hold. Suppose the PQ-learning algorithm is run with a step-size rule $\beta_t=\beta/(\lambda+t)$ with $\beta=2/c$ and $\lambda=16L/(c^2)$.  Then,
\begin{align*}
{\mathbb E}[\|\mathbf{T}Q_k-Q_{k,t}\|_D^2]\le\frac{512|{\cal S}||{\cal A}|}{(1-\gamma)^2}\frac{L}{c^3}\cdot\frac{1}{\lambda+t},\quad \forall t \ge 0.
\end{align*}
\end{proposition}
 Proposition~\ref{prop:inner-iteration} ensures that the inner iterate, $Q_{k,t}$, converges to the solution of the subproblem at the rate of ${\cal O}(1/t)$. Combining Proposition~\ref{prop:inner-iteration} with Proposition~\ref{prop:outer-iteration}, the overall sample complexity can be easily derived.
\begin{theorem}[Sample Complexity~I]\label{prop:sample-complexity}
Let $\beta_t=\beta/(\lambda+t)$ with $\beta=2/c$ and $\lambda=16L/(c^2)$ and the maximum number of steps
$$N_k = N \ge \frac{2048|{\cal S}||{\cal A}|}{\varepsilon^2(1-\gamma)^4}\frac{L}{c^3}.$$
PQ-learning achieves an $\varepsilon$-optimal solution, ${\mathbb E}[\|Q_T-Q^*\|_\infty]\le \varepsilon$,  with the number of samples at most
\begin{align*}
\frac{2048|{\cal S}||{\cal A}|}{\varepsilon^2(1-\gamma)^4\ln\gamma^{-1}}\frac{L}{c^3}\ln \left(\frac{4}{(1-\gamma)\varepsilon}\right).
\end{align*}
\end{theorem}

Note that in Theorem~\ref{prop:sample-complexity}, the constant $c$ and $L$ depend on $|{\cal S}|$ and $|{\cal A}|$. In particular, $c$ is upper bounded by $1/(|{\cal S}||{\cal A}|)$, while $L$ is lower bounded by $1/(|{\cal S}||{\cal A}|)$. The upper and lower bounds are achieved when the state-action distribution is uniform. In other words, the best sample complexity bound in Theorem~\ref{prop:sample-complexity} is obtained when the state-action distribution is uniform  because in this case, $c=L=1/(|{\cal S}||{\cal A}|)$ according to the definitions of $c$ and $L$, and the quantity, $\frac{L}{c^3}$, is minimized. Accordingly, the sample complexity becomes
$\tilde{\mathcal{O}}\left(\frac{|{\cal S}|^3|{\cal A}|^3}{\varepsilon^2(1-\gamma)^4}\right)$.
Since the per-iteration (inner iteration) complexity is ${\cal O}(1)$, the run-time complexity of PQ-learning has the same order as the sample complexity.

Although Theorem~\ref{prop:sample-complexity} provides a finite-time convergence analysis of PQ-learning in terms of the Q-function, it does not reflect convergence of the corresponding policy recovered from $Q_T$, namely, $\pi_{Q_T}(s):=\argmax_{a\in {\cal A}} Q_T(s,a)$. In the sequel, we focus on the convergence of the policy $\pi_{Q_T}$. A policy, $\pi_{Q_T}$, is called the $\varepsilon$-optimal policy if ${\mathbb E}[\|V^{\pi_{Q_T}} -V^*\|_\infty]\le\varepsilon$ holds. Before proceeding, we present a simple lemma that characterizes the relationship between the ${\mathbb E}[\|Q_T-Q^*\|_\infty]$ with ${\mathbb E}[\|V^{\pi_{Q_T}} -V^*\|_\infty]$.

\begin{lemma}\label{lem:relation} If ${\mathbb E}[\|Q_T-Q^*\|_\infty]\le\varepsilon$, then
${\mathbb E}[\|V^{\pi_{Q_T}}-V^*\|_\infty]\le\frac{2\varepsilon}{1-\gamma}.$
\end{lemma}
Invoking Lemma~\ref{lem:relation} and Theorem~\ref{prop:sample-complexity}, we immediately arrive at the following result.
\begin{theorem}[Sample Complexity~II]\label{prop:sample-complexity2}
Under the same setting as above, PQ-learning returns an $\varepsilon$-optimal policy $\pi_{Q_T}$ such that ${\mathbb E}[\|V^{\pi_{Q_T}}-V^*\|_\infty]\le \varepsilon$, with $\varepsilon \leq 1$, with the number of samples at most
 \begin{align*}
  \frac{8192|{\cal S}||{\cal A}|}{\varepsilon^2 (1-\gamma)^6 \ln\gamma^{-1}}\frac{L}{c^3}\ln \left( {\frac{8}{(1-\gamma)^2\varepsilon}}\right),
 \end{align*}
where $c$ and $L$ are constants as defined before.
\end{theorem}

Similarly, when the state-action distribution is uniform, the sample complexity bound for PQ-learning to obtain an $\varepsilon$-optimal policy becomes  $\tilde{\mathcal{O}}\left(\frac{|{\cal S}|^3|{\cal A}|^3}{\varepsilon^2(1-\gamma)^6}\right)$. It is worth pointing out that the sample complexity lower bound for solving DMDP under a generative sampling model established in the literature is $\Omega\left(\frac{|{\cal S}||{\cal A}|}{\varepsilon^2(1-\gamma)^3}\right)$~\citep{azar2013minimax}. Hence, PQ-learning is \emph{optimal} in terms of dependence on $\epsilon$, but not in the dependence on $|{\cal S}|, |{\cal A}|$ and the factor $1-\gamma$. Nonetheless, the sample complexity greatly improves over the existing results on standard Q-learning as reported in Table~\ref{table:comparison}, e.g.,  $\tilde{\cal O}( (|{\cal S}||{\cal A}|)^{\frac{2\ln(1/\varepsilon)}{1-\gamma}}/(\varepsilon^2))$ in~\citep{even2003learning}.


\section{Technical Proofs}
In this section, we provide detailed analysis of the main results described in the previous section. The characterizations of inner and outer iteration convergences, as stated in Propositions~\ref{prop:outer-iteration} and~\ref{prop:inner-iteration} form the backbone of the main results. Below we present the proofs for these two propositions. As can be seen, the analysis of PQ-learning is fairly simple and elegant based on the contraction property and standard arguments of SGD convergence.

\subsection{Proof of Proposition~\ref{prop:outer-iteration}}
We have
\begin{align*}
{\mathbb E}[\|Q_{k+1}-Q^*\|_\infty] \le& {\mathbb E}[\|Q_{k+1}-\mathbf{T}Q_{k}\|_\infty]+ {\mathbb E}[\|(\mathbf{T}Q_{k}-Q^*\|_\infty]\\
\le&\sqrt{{\mathbb E}[\|Q_{k+1}-\mathbf{T}Q_{k}\|_2^2]}+ {\mathbb E}[\|\mathbf{T}Q_{k}-\mathbf{T}Q^*\|_\infty]\\
\le&\sqrt{\varepsilon_{k+1}}+ \gamma {\mathbb E}[\|Q_k-Q^*\|_\infty],
\end{align*}
where the last inequality comes from the contraction property of the Bellman operator $\mathbf{T}$. Therefore, we have
\begin{align*}
&{\mathbb E}[\|Q_{k+1}-Q^*\|_\infty] \le\sqrt{\varepsilon_{k+1}}+\gamma {\mathbb E}[\|Q_k-Q^*\|_\infty].
\end{align*}
Combining the last inequality over $k=0,1,\ldots,T-1$, the desired result is obtained. $\quad \blacksquare$

\subsection{Proof of Proposition~\ref{prop:inner-iteration}}
Proposition~\ref{prop:inner-iteration} describes the convergence of the SGD subroutine when applied to solving the subproblem, $\min_{Q\in {\mathbb R}^{|{\cal S}||{\cal A}|}} l(Q;Q_k)$. Our proof follows the standard analysis in~\citet[Theorem~4.7]{bottou2018optimization}, adapted to the specific problem. Notice that the objective $l(Q;Q_k)$ is $c$-strongly convex and $L$-Lipschitz smooth.
\begin{lemma}[Strong convexity and Lipschitz continuity]\label{lemma:strong-convex-f-Lipschitz-continuity}
The objective function $l(Q;Q_k)$ is $c$-strongly convex with $c=\min_{s\in {\cal S},a\in {\cal A}} d_a(s)$ and is Lipschitz continuous with parameter $L=\max_{s\in {\cal S},a \in {\cal A}} d_a(s)$.
\end{lemma}
Moreover, we introduce two key lemmas showing that the variance of the stochastic gradient and target estimates can be properly bounded. Proofs are deferred in Appendix.
\begin{lemma}[Boundedness of variance]\label{lemma:basic-relation}
We have for any $Q\in {\mathbb R}^{|{\cal S}||{\cal A}|}$,
\begin{align*}
&{\mathbb E}[\|\tilde \nabla_Q l(Q;Q_{k})\|_2^2|Q_k,Q]\le 12 \gamma^2 |{\cal S}||{\cal A}| \|Q^*-Q_k\|_\infty^2+8\|\nabla_Q l(Q;Q_{k})\|_{D^{-1}}^2+\frac{18|{\cal S}||{\cal A}|}{(1-\gamma)^2}.
\end{align*}
\end{lemma}

\begin{lemma}[Boundedness of estimate]\label{lemma:basic-relation-2}
Suppose ${\mathbb E}[\|Q_i-\mathbf{T}Q_{i-1}\|_2^2]\le\varepsilon, \forall i\leq k$ and $\epsilon\leq (1-\gamma)^2$. Then $${\mathbb E}[\|Q_k-Q^*\|_\infty^2]\leq \frac{8}{(1-\gamma)^2}.$$
\end{lemma}

Invoking the smoothness and strong convexity conditions, one can easily show that
\begin{align*}
& \|\mathbf{T}Q_k-Q_{k,t+1}\|_D^2- \|\mathbf{T}Q_k-Q_{k,t}\|_D^2\\
\le& -2\beta_t \| D\mathbf{T}Q_k-DQ_{k,t}\|_2^2 + L\beta_t^2{\mathbb E}[\| \tilde \nabla_Q l(Q_{k,t};Q_{k})\|_2^2 |Q_{k,t},Q_k]\\
\le& (8L\beta_t^2-2c\beta_t) \|\mathbf{T}Q_k-Q_{k,t}\|_D^2+12|{\cal S}||{\cal A}|\gamma^2 L\beta_t^2 \|Q^*-Q_k\|_\infty^2+\frac{18|{\cal S}||{\cal A}|L\beta_t^2}{(1-\gamma)^2}.
\end{align*}
Recall that the step-size satisfies $0\le\beta_t\le \frac{c}{8L}$, which implies $8L\beta_t^2-2\beta_t c\le-c\beta_t$. Thus,
\begin{align*}
&{\mathbb E}[\|\mathbf{T}Q_k-Q_{k,t+1}\|_D^2]\le (1-c\beta_t) {\mathbb E}[\|\mathbf{T}Q_k-Q_{k,t}\|_D^2]+\beta_t^2\left(\frac{114|{\cal S}||{\cal A}|L}{(1-\gamma)^2}\right).
\end{align*}
By induction, it follows from the above recursion and $\beta_t=\frac{\beta}{\lambda+t}$ that
\begin{align*}
{\mathbb E}[\|\mathbf{T}Q_k-Q_{k,t}\|_D^2]\le\frac{v}{\lambda+t},\quad \forall t \ge 0, \forall k\geq 1
\end{align*}
 where $v =\max \{\lambda {\mathbb E}[\|\mathbf{T}Q_k  - Q_{k,0}\|_D^2],\beta^2C\}$, and $C=\frac{114|{\cal S}||{\cal A}|L}{{(1 - \gamma )^2 }}$. Note that the term $ {\mathbb E}[\|\mathbf{T}Q_k-Q_{k,0}\|_D^2]$ can be further bounded as
\begin{align*}
 {\mathbb E}[\|\mathbf{T}Q_k-Q_{k,0}\|_D^2]\le &\max_{s,a} d_a(s){\mathbb E}[\|\mathbf{T}Q_k-Q_{k,0}\|_2^2]\\
\le& 2L|{\cal S}||{\cal A}|{\mathbb E}[\|\mathbf{T}Q_k-\mathbf{T}Q^*\|_\infty^2]+2L|{\cal S}||{\cal A}|{\mathbb E}[\|\mathbf{T}Q^*-Q_k\|_\infty^2]\\
\le&2L|{\cal S}||{\cal A}|\gamma^2 {\mathbb E}[\|Q_k-Q^*\|_\infty^2]+2L|{\cal S}||{\cal A}| {\mathbb E}[\|Q^*-Q_k\|_\infty^2]\\
\le& \frac{32L|{\cal S}||{\cal A}|}{(1-\gamma)^2},
\end{align*}
Combing these facts then leads to the desired result in Proposition~\ref{prop:inner-iteration}. $\quad \blacksquare$

\section{Discussions}
In this paper, we introduce PQ-learning and provide its finite-time convergence and complexity analysis. The algorithm is relevantly simple, intuitive, and efficient. While this paper only focuses on the tabular case, the algorithm and analysis might be extended to PQ-learning with  function approximations. When linear function approximation is used,  a potential challenge arises from the mismatch between the $\infty$-norm used for the contraction property of the Bellman equation and the $2$-norm used for the projection of the Q-function onto the ranges of the feature vectors. Therefore, the composition of the Bellman operator and the projection operator is no longer a contraction. However, under certain conditions on the feature matrix, the composite mapping can be proven to be a contraction and our analysis would still apply. We leave this for future investigation.

Another important issue is the \emph{i.i.d.} assumption of samples. This assumption is rather restrictive as in practice the samples are often acquired from past trajectories or experiences. Our current result can be easily extended to Markovian sampling by applying standard mixing time arguments in stochastic optimization; see e.g.,~\cite{duchi2012ergodic,sun2018markov,dalal2018finite}. Note that our analysis of PQ-learning is mainly based on characterizing the cumulative errors from the SGD subroutines. Combining the existing result of SGD under Markovian samples with our analysis immediately leads to the finite-time convergence of PQ-learning under Markovian samples, which we leave for future investigation. Lastly, building on the stochastic optimization framework,  various methods such as the variance reduction techniques and acceleration schemes~\citep{kingma2014adam,allen2017katyusha,johnson2013accelerating,defazio2014saga,defazio2014finito} can be applied to replace the SGD subroutine used in the algorithm, to further improve the convergence or sample efficiency.

\bibliography{reference}

\clearpage
\newpage

\appendices

\begin{center}
{\Large \bf Appendix}
\end{center}

\section{Matrix representation}
 In this section, we introduce vector and matrix notations for the matrix representation of the Bellman operator. To this end, we introduce the following compact notations:
\begin{align*}
P:=& \begin{bmatrix}
   P_1\\
   \vdots\\
   P_{|{\cal A}|}\\
\end{bmatrix} \in {\mathbb R}^{|{\cal S}| \times |{\cal S}||{\cal A}|} ,\quad R:= \begin{bmatrix}
   R_1 \\
   \vdots \\
   R_{|{\cal A}|} \\
\end{bmatrix} \in {\mathbb R}^{|{\cal S}||{\cal A}|},\quad Q:= \begin{bmatrix}
   Q_1\\
  \vdots\\
   Q_{|{\cal A}|}\\
\end{bmatrix}\in {\mathbb R}^{|{\cal S}||{\cal A}|},\\
D_a:=& \begin{bmatrix}
   d_a(1) & & \\
   &  \ddots  &  \\
   & & d_a (|{\cal S}|)\\
\end{bmatrix}\in {\mathbb R}^{|{\cal S}| \times |{\cal S}|},\quad D:= \begin{bmatrix}
   D_1 & & \\
    &  \ddots  & \\
    & & D_{|{\cal A}|} \\
\end{bmatrix}\in {\mathbb R}^{|{\cal S}||{\cal A}| \times |{\cal S}||{\cal A}|},
\end{align*}
where $Q_a= Q(\cdot ,a) \in {\mathbb R}^{|{\cal S}|} ,a \in {\cal A}$ and $R_a(s):={\mathbb E}[r(s,a)|s,a]$. Note that $D$ is a nonsingular diagonal matrix with strictly positive diagonal elements. Moreover, for any deterministic policy, $\pi: {\cal S} \to {\cal A}$, we define the corresponding distribution vector
\begin{align*}
&\vec{\pi}(s):=e_{\pi(s)}\in \Delta_{|{\cal S}|},
\end{align*}
where $\Delta_{|{\cal S}|}$ is the set of all discrete probability distributions over ${\cal S}$, and define the matrix
\begin{align*}
\Pi_\pi: = \begin{bmatrix}
   \vec{\pi}(1)^T\otimes e_1^T\\
   \vec{\pi}(2)^T\otimes e_2^T\\
    \vdots\\
   \vec{\pi}(|{\cal S}|)^T \otimes e_{|{\cal S}|}^T\\
\end{bmatrix}\in {\mathbb R}^{|{\cal S}| \times |{\cal S}||{\cal A}|},
\end{align*}
where $e_j,j\in \{ 1,2,\ldots,|{\cal A}|\}$ is used to denote the $j$-th basis vector of dimension $|{\cal A}|$. Note that $\Pi_\pi$ is a matrix function which depends on the policy $\pi \in \Theta$. With these notations, the Bellman operator can be compactly written by
\begin{align*}
&\mathbf{T}Q=R+\gamma P\Pi_{\pi_Q} Q,
\end{align*}
where $\pi_Q(s):=\argmax_{a\in {\cal A}} e_s^T Q_a \in {\cal A}$. We note that, for any $\pi\in\Theta$, $P\Pi_{\pi}$ is the state-action pair transition probability matrix under the deterministic policy $\pi$, and $\Pi_{\pi_Q}$ is a nonlinear matrix function which depends on $Q$. This matrix representation of the Bellman operator plays an important role in the subsequent developments.

\section{Proofs of Technical Lemmas}

Throughout the remaining part, we denote by $F(Q):= l(Q;Q_k): = \frac{1}{2}\|R+P\Pi_{\pi_{Q_k}}Q_k-Q\|_D^2$.  Given a sample $(s,a,s')$, define the corresponding stochastic Bellman operator
\begin{align*}
{\bf\hat T}Q_k := \gamma\sum_{j\in {\cal S},i\in {\cal A}} {(e_i\otimes e_j )(e_{s'})^T \Pi_{\pi_{Q_k }}Q_k}+R
\end{align*}
and the matrix $\hat D:=(e_a\otimes e_s)(e_a\otimes e_s)^T$. Then, the stochastic gradient is written by
\begin{align*}
g(Q;Q_k)= -(\hat D{\bf\hat T}Q_k-\hat DQ) = -((e_a\otimes e_s)((e_{s'})^T \gamma\Pi_{\pi_{Q_k}}Q_k+(e_a\otimes e_s)^TR-(e_a\otimes e_s)^T Q)).
\end{align*}

\subsection{Proof of Lemma~\ref{lem:relation} }
We first show that if ${\mathbb E}[\|Q_T-Q^*\|_\infty]\le\varepsilon$, then
\begin{align*}
{\mathbb E}[\|Q^{\pi_{Q_T}}-Q^*\|_\infty]\le \frac{2\gamma\varepsilon}{1-\gamma}.
\end{align*}
This is because \begin{align*}
{\mathbb E}[\|Q^{\pi_{Q_T}}-Q^*\|_\infty]=&{\mathbb E}[\|Q^{\pi_{Q_T}}-\mathbf{T}Q_T+\mathbf{T}Q_T-Q^*\|_\infty]\\
\le& {\mathbb E}[\|Q^{\pi_T}-\mathbf{T}Q_T\|_\infty]+{\mathbb E}[\|\mathbf{T}Q_T-Q^*\|_\infty]\\
=& {\mathbb E}[\|R+\gamma P\Pi_{\pi_{Q_T}} Q^{\pi_{Q_T}}-(R+\gamma P\Pi_{\pi_{Q_T}} Q_T)\|_\infty]+\gamma\varepsilon\\
\le& {\mathbb E}[\gamma\|P\Pi_{\pi_{Q_T}}\|_\infty\|Q^{\pi_{Q_T}}-Q_T\|_\infty]+\gamma\varepsilon\\
\le& \gamma {\mathbb E}[\|Q^{\pi_{Q_T}}-Q_T\|_\infty]+\gamma\varepsilon\\
\le& \gamma {\mathbb E}[\|Q^{\pi_{Q_T}}-Q^*\|_\infty]+\gamma {\mathbb E}[\|Q^*-Q_T\|_\infty]+\gamma\varepsilon\\
\le& \gamma {\mathbb E}[\|Q^{\pi_{Q_T}}-Q^*\|_\infty]+2\gamma\varepsilon,
\end{align*}
where the second line is due to the triangle inequality, the second line uses the optimal Bellman equation, $Q^{\pi_{Q_T}} = R+\gamma P\Pi_{\pi_{Q_T}} Q^{\pi_{Q_T}}$ and $Q^* = \mathbf{T}Q^*$ and the contraction of the Bellman operator, the forth line is due to the hypothesis ${\mathbb E}[\|Q_T-Q^*\|_\infty]\le\varepsilon$, the fifth line is due to the Cauchy-Schwarz inequality, the sixth line is due to the fact that $P\Pi_{\pi_{Q_T}}$ is a stochastic matrix whose row sum is one, the seventh line is due to the triangle inequality, and the last line uses the hypothesis ${\mathbb E}[\|Q_T-Q^*\|_\infty]\le\varepsilon$. The desired result is obtained by rearranging the last inequality. Therefore, we have
\begin{align}
\|V^{\pi_{Q_T}}-V^*\|_\infty=&\max_{s \in {\cal S}} |V^{\pi_{Q_T}}(s)-V^*(s)|\nonumber\\
=&\max_{s\in {\cal S}} |Q^{\pi_{Q_T}}(s,\pi_{Q_T}(s))-\max_{a\in {\cal A}} Q^*(s,a)|\nonumber\\
\le&\max_{s\in {\cal S}} |Q^{\pi_{Q_T}}(s,\pi_T(s))-\max_{a \in {\cal A}}Q_T(s,a)|+\max_{s\in {\cal S}}|\max_{a\in {\cal A}} Q_T(s,a)-\max_{a\in {\cal A}} Q^*(s,a)|\nonumber\\
=& \max_{s\in {\cal S}}|Q^{\pi_{Q_T}}(s,\pi_{Q_T}(s))-\max_{a\in {\cal A}} Q_T (s,a)|+\|Q_T-Q^*\|_\infty.\label{eq:appendix:eq3}
\end{align}

For any fixed $s$, we have $|Q^{\pi_{Q_T}}(s,\pi_{Q_T}(s))-\max_{a\in {\cal A}}Q_T(s,a)|\le\max_{a\in {\cal A}} |Q_T(s,a)-Q^{\pi_{Q_T}}(s,a)|$. By taking the max over $s$ on the right-hand side first and then on the left-hand side, we obtain $\max_{s\in {\cal S}} |Q^{\pi_{Q_T}}(s,\pi_{Q_T}(s))-\max_{a\in {\cal A}}Q_T(s,a)|\le \|Q_T-Q^{\pi_{Q_T}}\|_\infty$. Using this inequality and taking the expectation on both sides of the inequality~\eqref{eq:appendix:eq3}, it is further bounded as
\begin{align*}
{\mathbb E}[\|V^{\pi_{Q_T}}-V^*\|_\infty]\le& {\mathbb E}[\|Q_T-Q^{\pi_{Q_T}}\|_\infty]+{\mathbb E}[\|Q_T-Q^*\|_\infty]\\
\le& {\mathbb E}[\|Q_T-Q^{\pi_{Q_T}}\|_\infty]+\varepsilon\\
\le& {\mathbb E}[\|Q_T-Q^*\|_\infty]+{\mathbb E}[\|Q^*-Q^{\pi_{Q_T}}\|_\infty]+\varepsilon\\
\le& {\mathbb E}[\|Q^*-Q^{\pi_{Q_T}}\|_\infty]+2\varepsilon\\
\le& \frac{2\gamma\varepsilon}{1-\gamma}+2\varepsilon\\
\le& \frac{2\varepsilon}{1-\gamma}.
\end{align*}
 This completes the proof. $\quad \blacksquare$

\subsection{Proof of Lemma~\ref{lemma:strong-convex-f-Lipschitz-continuity}}

Noting that
\begin{align*}
&F(Q)=\frac{1}{2}(R+P\Pi_{\pi_{Q_k}}Q_k)^T D(R+P\Pi_{\pi_{Q_k}}Q_k)+\frac{1}{2}Q^T DQ-(R+P\Pi_{\pi_{Q_k}}Q_k)^T DQ,
\end{align*}
which is a convex quadratic function. Since $F(Q)$ is twice differentiable, $F(Q)$ is $c$-strongly convex if and only if there exists a constant $c>0$ such that $\nabla_Q^2 F(Q)\ge cI$. Moreover, since the Hessian is $\nabla_Q^2 F(Q) = D$, and $D\ge\min_{s\in {\cal S},a\in {\cal A}}d_a(s)I$, $F(Q)$ is $c$-strongly convex with $c=\min_{s\in {\cal S},a\in {\cal A}} d_a(s)$.

Moreover, noting that $\nabla F(Q) = -D(R+P\Pi_{\pi_{Q_k}}Q_k-Q)$, we have
\begin{align*}
\|\nabla F(Q)-\nabla F(Q')\|_2=&\|D(R+P\Pi_{\pi_{Q_k}}Q_k-Q)-D(R+P\Pi_{\pi_{Q_k}}Q_k-Q')\|_2\\
=&\|D(Q-Q')\|_2\\
\le&\|D\|_2 \|Q-Q'\|_2,
\end{align*}
where $\|D\|_2=\sqrt {\lambda_{\max}(DD)}=\max_{s\in {\cal S},a\in {\cal A}}d_a(s)$, which proves the desired result. $\quad \blacksquare$

\subsection{Proof of Lemma~\ref{lemma:basic-relation}}
We want to show that
\begin{align*}
&{\mathbb E}[\|g(Q;Q_k)\|_2^2|Q_k,Q]\le 12 \gamma^2 |{\cal S}||{\cal A}| \|Q^*-Q_k\|_\infty^2+8\|\nabla F(Q)\|_{D^{-1}}^2+\frac{18|{\cal S}||{\cal A}|}{(1-\gamma)^2}.
\end{align*}
We first have
\begin{align*}
{\mathbb E}[\|g(Q;Q_k)\|_2^2|Q_k,Q]=&{\mathbb E}[\|\hat D{\bf\hat T}Q_k-\hat DQ |Q_k,Q]\\
=&{\mathbb E}[\| \hat D{\bf\hat T}Q_k- \hat DQ -(\hat D{\bf\hat T}Q^*-\hat DQ^*)+ \hat D{\bf\hat T}Q^*-\hat DQ^*\|_2^2|Q_k,Q]\\
\leq & 2{\mathbb E}[\|(e_a\otimes e_s)((e_{s'})^T \gamma\Pi_{\pi_{Q_k}}Q_k-(e_{s'})^T\gamma\Pi_{\pi_{Q^*}}Q^*)\\
&-(e_a\otimes e_s )((e_a\otimes e_s)^T Q - (e_a\otimes e_s)^T Q^*)\|_2^2|Q_k]+2\sigma^2,
\end{align*}
where $\sigma^2:={\mathbb E}[\|\hat D{\bf\hat T}Q^*-\hat D Q^*\|_2^2]$ and the last inequality is due to $\|a+b\|^2\le 2\|a\|^2+2\|b\|^2$. Now, we further have
\begin{align*}
&{\mathbb E}[\|g(Q;Q_k)\|_2^2 |Q_k,Q]\\
\le& 2{\mathbb E}[\| e_a\otimes e_s\|_2^2] {\mathbb E}[\|(e_{s'})^T (\gamma\Pi_{\pi_{Q_k}}Q_k-\gamma\Pi_{\pi_{Q^*}}Q^*)-(e_a\otimes e_s)^T (Q-Q^*)\|_2^2|Q_k,Q]+2\sigma^2\\
\le& 2{\mathbb E}[\|(e_{s'})^T (\gamma\Pi_{\pi_{Q_k}}Q_k-\gamma\Pi_{\pi_{Q^*}} Q^*)-(e_a\otimes e_s)^T (Q-Q^*)\|_2^2|Q_k,Q]+2\sigma^2\\
\le& 4\gamma^2 {\mathbb E}[\|(e_{s'})^T (\Pi_{\pi_{Q_k}}Q_k-\Pi_{\pi_{Q^*}}Q^*)\|_2^2|Q_k]+4{\mathbb E}[\|(e_a\otimes e_s)^T (Q-Q^*)\|_2^2|Q]+2\sigma^2\\
=&4\gamma^2 {\mathbb E}[(\Pi_{\pi_{Q_k}}Q_k-\Pi_{\pi_{Q^*}} Q^*)^T e_{s'}(e_{s'})^T (\Pi_{\pi_{Q_k}}Q_k-\Pi_{\pi_{Q^*}}Q^*)|Q_k,Q]\\
&+4{\mathbb E}[(Q-Q^*)^T (e_a\otimes e_s)(e_a\otimes e_s)^T (Q-Q^*)|Q]+2\sigma^2\\
=& 4\gamma^2\|\Pi_{\pi_{Q_k}}Q_k-\Pi_{\pi_{Q^*}}Q^*\|_{{\mathbb E}[e_{s'}(e_{s'})^T]}^2+4\|Q-Q^*\|_D^2+2\sigma^2\\
\le& 4\gamma^2 |{\cal S}| \|\Pi_{\pi_{Q_k}}Q_k-\Pi_{\pi_{Q^*}} Q^*\|_\infty^2+4\|Q-Q^*\|_D^2+2\sigma^2,
\end{align*}
where the first inequality follows from the Cauchy–Schwarz inequality and Holder's inequality, the second inequality is due to ${\mathbb E}[\|e_a\otimes e_s\|_2^2]\le 1$, and the third inequality is due to $\|a+b\|^2\le 2\|a\|^2+2\|b\|^2$. Invoking the following inequality
\begin{align*}
\|\Pi_{\pi_{Q_k}}Q_k-\Pi_{\pi_{Q^*}}Q^*\|_\infty^2\le \|Q_k-Q^*\|_\infty^2,
\end{align*}
we further bound ${\mathbb E}[\|g(Q;Q_k)\|_2^2|Q_k,Q]$ as
\begin{align*}
{\mathbb E}[\|g(Q;Q_k)\|_2^2|Q_k,Q] \le 4\gamma^2 |{\cal S}|\|Q_k-Q^*\|_\infty^2+4\|Q-Q^*\|_D^2+2\sigma^2.
\end{align*}

For later uses, we will bound $\|Q-Q^*\|_D^2$ in terms of $\nabla F(Q)=-(D\mathbf{T}Q_k-DQ)$. To this end, we first observe
\begin{align*}
{\mathbb E}[\|g(Q;Q_k)\|_2^2|Q_k,Q]\le& 4\gamma^2 |{\cal S}| \|Q_k-Q^*\|_\infty^2+4\|Q-Q^*\|_D^2+2\sigma^2\\
=& 4\gamma^2|{\cal S}|\|Q_k-Q^*\|_\infty ^2 + 4\|DQ-DQ^*\|_{D^{-1}}^2+2\sigma^2\\
=&4\gamma^2 |{\cal S}|\|Q_k-Q^*\|_\infty^2+4 \|DQ^*-DQ-\nabla F(Q)+\nabla F(Q)\|_{D^{-1}}^2+2\sigma^2\\
\le& 4\gamma^2 |{\cal S}|\|Q_k-Q^*\|_\infty^2+8\|DQ^*-DQ-\nabla F(Q)\|_{D^{-1}}^2+8\|\nabla F(Q)\|_{D^{-1}}^2+2\sigma^2\\
=& 4\gamma^2 |{\cal S}|\|Q_k-Q^*\|_\infty^2+8\|DQ^*-D\mathbf{T}Q_k\|_{D^{-1}}^2+8\|\nabla F(Q)\|_{D^{-1}}^2+2\sigma^2\\
\le& 4\gamma^2 |{\cal S}|\|Q_k-Q^*\|_\infty^2+8L\|Q^*-\mathbf{T}Q_k\|_2^2+8\|\nabla F(Q)\|_{D^{-1}}^2+2\sigma^2\\
\le& 4\gamma^2 |{\cal S}|\|Q_k-Q^*\|_\infty^2+8L|{\cal S}||{\cal A}| \|\mathbf{T}Q^*-\mathbf{T}Q_k\|_\infty^2+8\|\nabla F(Q)\|_{D^{-1}}^2+2\sigma^2\\
\le& 4\gamma^2 |{\cal S}| \|Q_k-Q^*\|_\infty^2+8|{\cal S}||{\cal A}|\gamma^2 \|Q^*-Q_k\|_\infty^2+8\|\nabla F(Q)\|_{D^{-1}}^2+2\sigma^2\\
\le& 12\gamma^2 |{\cal S}||{\cal A}|\|Q^*-Q_k\|_\infty^2+8\|\nabla F(Q)\|_{D^{-1}}^2+2\sigma^2,
\end{align*}
where the fourth line follows from $\|a+b\|^2\le 2\|a\|^2+2\|b\|^2$, the eight line comes from the contraction property of the Bellman operator, and the last line follows after simplifications.

Finally, we will find a bound on $\sigma^2$. By the definition, we have
\begin{align*}
\sigma^2=& {\mathbb E}[\|\hat D{\bf\hat T}Q^*- \hat D Q^*\|_2^2]\\
=& {\mathbb E}[\|(e_a\otimes e_s)((e_{s'})^T \gamma\Pi_{\pi_{Q^*}}Q^*-(e_a\otimes e_s)^TR-(e_a\otimes e_s)^T Q^*)\|_2^2]\\
\le& {\mathbb E}[\|(e_a\otimes e_s)\|_2^2] {\mathbb E}[\|(e_{s'})^T\gamma\Pi_{\pi_{Q^*}}Q^*-(e_a\otimes e_s)^TR-(e_a\otimes e_s)^TQ^*\|_2^2]\\
\le& {\mathbb E}[\|(e_{s'})^T \gamma\Pi_{\pi_{Q^*}}Q^*-(e_a\otimes e_s)^TR-(e_a\otimes e_s)^T Q^*\|_2^2]\\
\le& 3{\mathbb E}[\|(e_{s'})^T\gamma\Pi_{\pi_{Q^*}} Q^*\|_2^2]+3{\mathbb E}[\|(e_a\otimes e_s)^TR\|_2^2]+3{\mathbb E}[\|(e_a\otimes e_s)^TQ^*\|_2^2]\\
\le& 3{\mathbb E}[\|(e_{s'})^T \gamma\Pi_{\pi_{Q^*}}\|_2^2 \|Q^*\|_2^2]+3{\mathbb E}[\|(e_a\otimes e_s)^TR\|_2^2]+3{\mathbb E}[\|e_a  \otimes e_s\|_2^2 \|Q^*\|_2^2]\\
\le& 3|{\cal S}||{\cal A}|{\mathbb E}[\|(e_{s'})^T \gamma\Pi_{\pi_{Q^*}}\|_2^2 \|Q^*\|_\infty^2]+3{\mathbb E}[\|(e_a\otimes e_s)^T R\|_2^2]+3|{\cal S}||{\cal A}|{\mathbb E}[\|e_a\otimes e_s\|_2^2 \|Q^*\|_\infty ^2]\\
\le& 3|{\cal S}||{\cal A}|\|Q^*\|_\infty^2+3+3|{\cal S}||{\cal A}|\|Q^*\|_\infty^2\\
\le& 3|{\cal S}|{\cal A}|\frac{1}{(1-\gamma)^2} + 3 + 3|{\cal S}||{\cal A}|\frac{1}{(1-\gamma)^2}\\
\le& \frac{9|{\cal S}||{\cal A}|}{(1-\gamma)^2},
\end{align*}
where third line is due to the Cauchy–Schwarz inequality and Holder's inequality, the forth line is due to ${\mathbb E}[\|e_a\otimes e_s\|_2^2]\le 1$, the fifth line uses $\|a+b+c\|^2\le 3\|a\|^2+3\|b\|^2+3\|c\|^2$, the sixth line follows from the Cauchy–Schwarz inequality again, the eighth line comes from $\|(e_{s'})^T \gamma\Pi_{\pi_{Q^*}}\|_2^2\le 1,\|(e_a\otimes e_s)^TR\|_2^2\le 1,\|e_a  \otimes e_s\|_2^2\le 1$, the ninth line is due to $\|Q^*\|_\infty\le 1/(1-\gamma)$, and tenth line follows after simplifications. Combining the bound $\sigma^2\le 9|{\cal S}||{\cal A}|/(1-\gamma)^2$ with the previous result, the desired result follows. This completes the proof. $\quad \blacksquare$

\subsection{Proof of Lemma~\ref{lemma:basic-relation-2}}
We want to show that if ${\mathbb E}[\|Q_i-\mathbf{T}Q_{i-1}\|_2^2]\le\varepsilon, \forall i\leq k$ and $\epsilon\leq (1-\gamma)^2$, then ${\mathbb E}[\|Q_k-Q^*\|_\infty^2]\leq \frac{8}{(1-\gamma)^2}$.  We start with the following claim.

\emph{Claim:} If ${\mathbb E}[\|Q_i- \mathbf{T}Q_{i-1}\|_2^2]\le\varepsilon,i\in \{1,\ldots ,T\}$ with $T\geq 1$, then we have
\begin{align*}
&{\mathbb E}[\|Q_T-Q^*\|_\infty^2]\le\left(\frac{1+\gamma ^2}{1-\gamma^2}\right)\varepsilon \frac{2}{1-\gamma^2}+\left(\frac{\gamma^2+1}{2}\right)^T {\mathbb E}[\|Q_0-Q^*\|_\infty^2].
\end{align*}
This is because
\begin{align*}
{\mathbb E}[\|Q_T-Q^*\|_\infty^2]=& {\mathbb E}[\|Q_T-\mathbf{T}Q_{T-1}+\mathbf{T}Q_{T-1}-Q^*\|_\infty^2]\\
\le& {\mathbb E}[(\|Q_T-\mathbf{T}Q_{T-1}\|_\infty+\|\mathbf{T}Q_{T-1}-Q^*\|_\infty)^2]\\
\le& {\mathbb E}[(1+\delta^{-1})\|Q_T- \mathbf{T}Q_{T-1}\|_\infty^2+(1+\delta) \|\mathbf{T}Q_{T-1}-Q^*\|_\infty^2]\\
=&(1+\delta^{-1}) {\mathbb E}[\|Q_T-\mathbf{T}Q_{T-1}\|_2^2]+(1+\delta){\mathbb E}[\|\mathbf{T}Q_{T-1}-\mathbf{T}Q^*\|_\infty^2]\\
\le& (1+\delta^{-1})\varepsilon+(1+\delta)\gamma^2 {\mathbb E}[\|Q_{T-1}-Q^*\|_\infty^2],
\end{align*}
where the second line is due to the triangle inequality, the third line is due to the fact that $\|a+b\|_2^2 \le (1+\delta)\|a\|_2^2 + (1+\delta^{-1})\|b\|_2^2$ for any $\delta >0$, and the last line is due to the hypothesis, ${\mathbb E}[\|Q_i- \mathbf{T}Q_{i-1}\|_2^2]\le\varepsilon,i\in \{1,\ldots ,T\}$. Note that we can choose $\delta>0$ such that $(1+\delta)\gamma^2<1$ or equivalently, $\delta<\frac{1-\gamma^2}{\gamma^2}$. Simply choosing $\delta=\frac{1-\gamma^2}{2\gamma^2}$ yields
\begin{align*}
{\mathbb E}[\|Q_T-Q^*\|_\infty^2]\le\left(1+\frac{2\gamma^2}{1-\gamma^2}\right)\varepsilon+\frac{\gamma^2+1}{2}{\mathbb E}[\|Q_{T-1}-Q^*\|_\infty^2].
\end{align*}

By the induction argument in $T$, we get
\begin{align*}
{\mathbb E}[\|Q_T-Q^*\|_\infty^2]\le& \left(1+\frac{2\gamma^2}{1-\gamma^2}\right)\varepsilon \sum_{k=1}^T\left(\frac{\gamma^2+1}{2} \right)^k+\left(\frac{\gamma^2+1}{2}\right)^T {\mathbb E}[\|Q_0-Q^*\|_\infty^2]\\
\le& \left(1+\frac{2\gamma^2}{1-\gamma^2}\right)\varepsilon \sum_{k=1}^\infty \left(\frac{\gamma^2+1}{2}\right)^k +\left(\frac{\gamma^2 +1}{2}\right)^T {\mathbb E}[\|Q_0-Q^*\|_\infty^2]\\
=& \left(\frac{1+\gamma^2}{1-\gamma^2} \right)\varepsilon \frac{2}{1-\gamma^2}+\left( \frac{\gamma^2+1}{2}\right)^T {\mathbb E}[\|Q_0-Q^*\|_\infty^2],
\end{align*}
which proves the claim. As an immediate result,

\begin{align}
{\mathbb E}[\|Q_k-Q^*\|_\infty^2]\le&\left(\frac{1+\gamma^2}{1-\gamma^2}\right)\varepsilon \frac{2}{1-\gamma^2}+\left(\frac{\gamma^2+1}{2}\right)^k {\mathbb E}[\|Q_0-Q^*\|_\infty^2]\nonumber\\
\le& \left(\frac{1+\gamma^2}{1-\gamma^2}\right)\varepsilon \frac{2}{1-\gamma^2} + \frac{4}{(1-\gamma)^2}\nonumber\\
\le& \frac{2}{(1-\gamma)^2}\varepsilon \frac{2}{1-\gamma^2}+\frac{4}{(1-\gamma)^2}\nonumber\\
\le& \frac{4}{1-\gamma^2} + \frac{4}{(1-\gamma)^2},\nonumber\\
\le& \frac{8}{(1-\gamma)^2},\label{eq:appendix:eq1}
\end{align}
where the first line comes from the claim, the second inequality is due to ${\mathbb E}[\|Q_0-Q^*\|_\infty^2] \le 1/((1-\gamma)^2)$ and $(\gamma^2+1)/2<1$, the third line comes from $\frac{1+\gamma^2}{1-\gamma^2}\le\frac{2}{(1-\gamma)^2}$, the fourth line follows from the hypothesis $\varepsilon\le(1-\gamma)^2$, and the last line follows after simplifications. For the case $k=0$, the bound is also valid since ${\mathbb E}[\|Q_0-Q^*\|_\infty^2]\le \frac{4}{(1-\gamma)^2}$. Taking the total expectation on both sides  and combining it with~\eqref{eq:appendix:eq1}, we arrive at the desired conclusion. $\quad \blacksquare$

\section{Proof of Theorem~\ref{prop:sample-complexity}}

The main line of the proof is to balance the inner and outer iteration numbers to achieve ${\mathbb E}[\|Q_T-Q^*\|_\infty]\le\varepsilon$.

First of all, we define numbers $\xi_i >0,i\in \{1,2,\ldots,T\}$ as those satisfying ${\mathbb E}[\|Q_i-\mathbf{T}Q_{i-1}\|_\infty]\le\sqrt{\xi_i},i \in \{1,2,\ldots ,T\}$ through the previous inner iterations. By Proposition~\ref{prop:outer-iteration}, we first have
\begin{align*}
{\mathbb E}[\|Q_T-Q^*\|_\infty]\le&\sum_{k=1}^T \gamma^{T-k}\sqrt{\xi_k}+\gamma^T {\mathbb E}[\|Q_0-Q^*\|_\infty]\le\sum_{k=1}^T \gamma^{T-k}\sqrt{\xi_k}+\frac{2\gamma^T}{1-\gamma},
\end{align*}
where the last inequality is due to ${\mathbb E}[\|Q_0-Q^*\|_\infty]\le 2/(1-\gamma)$ and ${\mathbb E}[\|Q_i-\mathbf{T}Q_{i-1}\|_\infty]\le\sqrt{\xi_i},i \in \{1,2,\ldots ,T\}$. Moreover, by Proposition~\ref{prop:inner-iteration}, a constant $N_k=N$ implies that $\xi_i$ can be uniformly bounded by a constant $\xi$. Letting $\xi_i = \xi$, the last inequality can be bounded as
\begin{align*}
{\mathbb E}[\|Q_T-Q^*\|_\infty]\le \frac{\sqrt\xi}{1-\gamma}+\frac{2\gamma^T}{1-\gamma}.
\end{align*}

To achieve ${\mathbb E}[\|Q_T-Q^*\|_\infty]\le\varepsilon$, it suffices to ensure
\begin{align*}
&{\rm Inner\,\,iteration}:\frac{\sqrt\xi}{1-\gamma}\le\frac{1}{2}\varepsilon,\\
&{\rm Outer\,\,iteration}:\frac{2\gamma^T}{1-\gamma}\le\frac{1}{2}\varepsilon.
\end{align*}
 By rearranging terms and taking the logarithm on both sides of the second inequality, it follows that the second inequality is equivalent to
\begin{align*}
T\ge\ln\left(\frac{1-\gamma}{4}\varepsilon\right)/\ln\gamma =\ln\left(\frac{4}{(1-\gamma)\varepsilon}\right)/\ln\gamma^{-1}.
\end{align*}
The first inequality is equivalent to $\xi\le \frac{\varepsilon^2(1-\gamma)^2}{4}$, which is ensured if
\begin{align}
{\mathbb E}[\|Q_i-\mathbf{T}Q_{i-1}\|_\infty]\le\sqrt \frac{\varepsilon^2(1-\gamma)^2}{4},\quad \forall i \in \{ 1,2, \ldots ,T\}.\label{eq:appendix:eq2}
\end{align}

The  square of the left-hand side of the above inequality is bounded by
\begin{align*}
{\mathbb E}[\|Q_i-\mathbf{T}Q_{i-1}\|_\infty]^2\le& {\mathbb E}[\|Q_i-\mathbf{T}Q_{i-1}\|_2]^2\\
\le& {\mathbb E}[\|Q_i- \mathbf{T}Q_{i-1}\|_2^2]\\
\le& c^{-1} {\mathbb E}[\|Q_i-\mathbf{T}Q_{i-1}\|_D^2]\\
\le& \frac{1}{\lambda+N}\frac{512|{\cal S}||{\cal A}|}{(1-\gamma)^2}\frac{L}{c^3},
\end{align*}
where the second line is due to the Jensen's inequality and the last line is due to Theorem~\ref{prop:sample-complexity}. Therefore, a sufficient condition to ensure~\eqref{eq:appendix:eq2} is
$\frac{1}{\lambda+N}\frac{512|{\cal S}||{\cal A}|}{(1-\gamma)^2}\frac{L}{c^3} \le \frac{\varepsilon^2(1-\gamma)^2}{4}.$
The last inequality is implied by
\begin{align*}
\frac{2048|{\cal S}||{\cal A}|}{\varepsilon^2(1-\gamma)^4}\frac{L}{c^3}\le N.
\end{align*}
 In summary, if the total number of inner iteration $N$ is greater than or equal to $\frac{2048|{\cal S}||{\cal A}|}{\varepsilon^2(1-\gamma)^4}\frac{L}{c^3}$, then the first inequality $\frac{\sqrt\xi}{1-\gamma}\le\frac{1}{2}\varepsilon$ holds. By combining the total number of inner iterations and outer iterations, the total number of samples that are required to achieve ${\mathbb E}[\|Q_T-Q^*\|_\infty]\le\varepsilon$ is $\frac{2048|{\cal S}||{\cal A}|}{\varepsilon^2(1-\gamma)^4 \ln\gamma^{-1}}\frac{L}{c^3}\ln\left(\frac{4}{(1-\gamma)\varepsilon}\right)$. This completes the proof. $\quad \blacksquare$




\end{document}